\documentclass[letterpaper,10 pt,conference]{IEEEtran}
\IEEEoverridecommandlockouts
\usepackage{cite}
\usepackage{amsmath,amssymb,amsfonts}
\usepackage{algorithmic}
\usepackage{graphicx}
\usepackage{textcomp}
\usepackage{epstopdf}
\usepackage{xcolor}
\usepackage{mathrsfs}

\def\BibTeX{{\rm B\kern-.05em{\sc i\kern-.025em b}\kern-.08em
    T\kern-.1667em\lower.7ex\hbox{E}\kern-.125emX}}
\begin{document}

\title{\LARGE \bf Swarm of One: Bottom-up Emergence of Stable Robot Bodies from Identical Cells\\

\thanks{$^{1}$Trevor Smith is a National Science Foundation Graduate Research Fellow with the Department of Mechanical and Aerospace engineering at WVU$^\dagger$
        {\tt\small trs0024@mix.wvu.edu}}%
\thanks{$^{2}$R. Michael Butts is a Master's student with the Lane Department of Computer Science and Electrical Engineering at WVU$^\dagger$
        {\tt\small rmb0034@mix.wvu.edu}}%
\thanks{$^{3}$Nathan Adkins is an Undergraduate student with the Lane Department of Computer Science and Electrical Engineering at WVU$^\dagger$
        {\tt\small npa00003@mix.wvu.edu}}%
\thanks{$^{4}$Yu Gu is faculty with the Department of Mechanical and Aerospace engineering at WVU$^\dagger$
        {\tt\small yu.gu@mail.wvu.edu}}%
\thanks{$^{\dagger}$ West Virginia University, Morgantown, WV 26506, USA}%

\thanks{$*$ Authors Trevor Smith and R. Michael Butts contributed equally to this work.}

\thanks{This study was supported in part by the National Science Foundation Award \#1851815 and the National Science Foundation Graduate Research Fellowship Award \#2136524.}
}

\author{ $^*$Trevor Smith$^{1}$, $^*$R. Michael Butts$^{2}$, Nathan Adkins$^{3}$, Yu Gu$^{4}$ \\
\textit{West Virginia University}\\
Morgantown, USA 
}

\maketitle

\begin{abstract}
 Unlike most human-engineered systems, biological systems are emergent from low-level interactions, allowing much broader diversity and superior adaptation to the complex environments.  Inspired by the process of morphogenesis in nature, a bottom-up design approach for robot morphology is proposed to treat a robot's body as an emergent response to underlying processes rather than a predefined shape. This paper presents Loopy, a ``Swarm-of-One" polymorphic robot testbed that can be viewed simultaneously as a robotic swarm and a single robot. Loopy’s shape is determined jointly by self-organization and morphological computing using physically linked homogeneous cells. Experimental results show that Loopy can form symmetric shapes consisting of lobes. Using the the same set of parameters, even small amounts of initial noise can change the number of lobes formed. However, once in a stable configuration, Loopy has an ``inertia" to transfiguring in response to dynamic parameters. By making the connections among self-organization, morphological computing, and robot design, this paper lays the foundation for more  adaptable robot designs in the future.
 
\end{abstract}

%================================================================== INTRODUCTION
\section{Introduction}
% Intro Paragraph
Most robots today were designed through a primarily top-down process, relying heavily on human creativity to determine the form and behavior of a robot to meet a set of design requirements. This work takes an alternative path in designing a single robot's morphology from the bottom-up interactions among decentralized agents. Inspired by morphogenesis (how cellular patterns and shapes emerge in nature) \cite{davies2020engineering,goryachev2020patterning}, we see a robot’s body as an emergent response to the underlying processes instead of a predefined shape.

%------------------------------------------------ cool loopy picture
% faded transient response from 1 initial condition
\begin{figure}[htbp]
\centerline{\includegraphics[scale = .275]{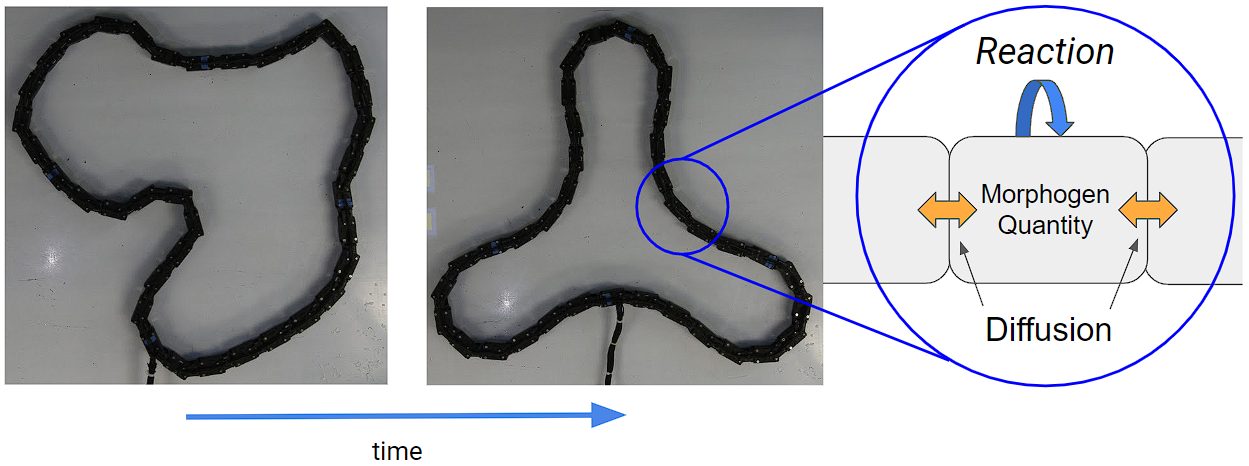}}
\caption{ Loopy transfiguring into a stable emergent body from a random initial condition (left). Loopy's shape (middle) is generated by an intra-cellular reaction-diffusion system (right) and the morphological computation of the inter-cellular physical links}
\label{coolLoopy}
\end{figure}
% -----------------------------------------------------------------
We developed a new kind of robot, named Loopy, to explore bottom-up, self-organization-based, design methods. Loopy is a single robot consisting of a closed chain of many rotary actuators acting as independent one-degree-of-freedom (1-DoF) agents (each loosely analogous to a biological cell). These agents locally interact and make decisions, forming a robot swarm. We call Loopy a ``Swarm-of-One" because it can be viewed simultaneously as a single robot and a swarm of simple robots, similar to multi-cellular organisms (Fig. \ref{coolLoopy}). Potential future applications of this morphogenic approach would be reconfigurable robots that adapt their shape to new environments, allowing for a greater understanding of the connection between biological and robotic morphology.

    % Bottom-up Paragraph
    When designing a robot, the design space (i.e., the space of all possible designs) is often infinite dimensional, making it challenging to find acceptable solutions, let alone to optimize one, unless the degrees of freedom in design are severely constrained \cite{kang2010approach,nardi2019practical}. Despite extensive research on design methodology \cite{gericke2011comparisons,tomiyama2009design,kapurch2010nasa}, human experiences and creativity are still essential in finding useful designs, making the robot design process today arguably more of an art than a science. Products of this mostly predictive, or top-down, approach of robot design often fail to adapt to the complexity posed by the real world \cite{pollack2000evolutionary}. Viewed from the other direction, the bottom-up forces used by nature to shape the world we are living in is a conceivable but difficult direction to emulate. Past research inspired by nature often followed an evolutionary path \cite{floreano2000evolutionary,cheney2014unshackling,wang2019neural} to design modular robots. Self-organization, another key contributor to the wonderful diversity of nature’s creations \cite{pfeifer2007self}, has not been adequately explored for solving bottom-up robot design problems, especially for single robots. When seeing both robot morphology and behavior as emergent through bottom-up interactions, the lines between them become blurred, along with the distinctions between the robot’s hardware and software \cite{castle2019virtual,crespi2008top}. 

    %Swarm Paragraph
    Recent advancements in self-assembling robotic systems have explored the formation of complex shapes through low-level interactions \cite{liu2022smores,abdel2022self,rubenstein2014programmable,o2010self,wei2010sambot,gross2006autonomous}. These systems often consist of a set of identical robots that are instructed to self-assemble/organize to a human-defined shape using a shape decomposer\cite{covell2022top}. This style of system addresses the problem of bottom-up shape formation of a top-down design. Similarly, formation control has demonstrated bottom-up control techniques to produce multi-agent formations based on predetermined geometry, as seen in \cite{oh2015survey,ahn2020formation}. On the other hand, emergent (i.e., not pre-determined) shape formation has been demonstrated within robot swarms, with the most common example being the bio-inspired flocking behavior following Boids rules \cite{reynolds1987flocks,hartman2006autonomous,alaliyat2014optimisation}, where a flock formation emerges from the local interactions of each agent.  

    More recently, biologically inspired bottom-up interactions have also been analyzed through the lens of morphogenesis, where simulated low-level chemical interaction between cells produce global emergent formations\cite{mamei2004experiments,sayama2010robust,jin2010morphogenetic}. Within these works, a common methodology is the use of Turing patterns, where a reaction of two chemicals (i.e., activator and inhibitor) are modeled as a set of partial differential equations \cite{turing1990chemical}. Subsequent works have extended this methodology to describe the formation of animals' limbs and fingers, as well as patterns on seashells and feathers, etc. \cite{marcon2012turing,nakamasu2009interactions,boettiger2009neural}. A common type of reaction diffusion model is the Fitzhugh-Nagumo equations, which produce stable sinusoidal distributions \cite{fitzhugh1955mathematical}. 
    
    In the work by Slavkov et. al \cite{slavkov2018morphogenesis}, Turing patterns have been applied to the Kilobot platform to produce emergent artificial morphogenesis. Due to the use of free and physically isolated agents, top-down designed cohesion rules are required to produce these formations. This differs from biological systems, where the cells are physically attached, exerting forces on each other, and taking advantage of morphological computing \cite{pfeifer2006morphological,fuchslin2013morphological,hauser2014morphological,gu2018cataglyphis} as a contributing factor in their shape creation \cite{ben2023morphological}. Building off this insight, this study contributes to the literature in the following ways.
\begin{itemize}
    \item Inspired by the field of morphogenesis, this paper makes the connection between self-organization and the bottom-up design of robot morphology. A robot’s body is treated as an emergent behavior instead of a deterministic pre-planned shape;
    \item We introduce Loopy, a ``Swarm-of-One", which can be viewed simultaneously as a robot swarm and a single robot, as a standard testbed for bottom-up robot design and control;
    \item With physically linked homogeneous cells, Loopy's shape is determined jointly by self-organization and morphological computing;  
    \item Experiments demonstrated Loopy’s ability to generate a variety of stable emergent shapes influenced by Loopy's inter-cellular environment (e.g., initial conditions and the closed-chain kinematic constraints)
\end{itemize}
The rest of this paper is outlined as follows. Section \ref{methodology} details the methodology utilized in this work. Section \ref{RaD} describes the results and discussion of the experiments. Finally, Section \ref{CaF} concludes and expresses the future work of this study.

%================================================================== METHODOLOGY
\section{Methodology}
    \label{methodology}
    
    \subsection{Problem Statement}
    The objective of this study is to find bottom-up robot morphology design approaches through the lenses of self-organization and morphological computation. This includes three components: 1. finding a general-purpose multi-cellular robot arrangement that can be configured into different shapes; 2. achieving stable emergent body shapes from random initial conditions using this testbed; 3. studying the properties of the emergent robot bodies during both transient and steady-state conditions.
    
    In this work, we assume the robot is made of a large number of homogeneous cells that are physically linked. Decisions are made locally and at no point, any of the cells in the system are made aware of the (either desired or current) global shape of the robot. Direct communication between cells can only happen “on the surface” (i.e., between immediate neighbors). This study also only focuses on generating 2D shapes.
    
 \subsection{Loopy System Design}
 \label{section:loopyDesign}
    The Loopy robot was designed as a polymorphic test platform for bottom-up design/control methods where conventional centralized methods are impractical. Embodied by a closed chain of $N$ identical rotary motors, without an orientation or global coordinate frame \cite{covell2022top}, Loopy is difficult to explicitly model macroscopically but simple microscopically. 
    
   Since loopy is a ring and not an open chain of motors-cells, such as a snake robot, the inter-cellular linkages enforce additional physical constraints. The first constraint is that neighboring cells are physically connected to each other, therefore Loopy can be described as a set of vectors whose sum has zero net displacement as shown in:

    \begin{equation}
        \sum \vec{s}_m = \vec{0}
        \label{sum_vectors_eq}
    \end{equation}
    
    \begin{equation}
         \vec{s}_m = se^{i(\theta_m + \theta_{m-1})}
        \label{vector_eq}
    \end{equation}
    
    Where $s$ is the length of each cell, $\theta_m$ is the external angle of the $m^{th}$ cell. Note that Loopy cells do not, in actuality, possess a head or tail, and the index is only used to aid in the description of the macroscopic model. In addition, this closed-loop constraint also makes Loopy a polygon, and thus the sum of all external angles must be equal to $2\pi$.  

    \begin{equation}
         \sum \theta_m = 2\pi
        \label{polygonSumAngles}
    \end{equation}
    
    Since Loopy is 2D, the cells cannot intersect each other. This constraint is expressed in \eqref{interPt} and \eqref{interSet} which evaluate if a pair of Loopy's cells intersect. Where $m$ and $n$ represent the $m^{th}$ and $n^{th}$ cells, respectively, $x$ is the global x-coordinate of the cell and $x_{intersect}$ is the global x-coordinate where the lines formed by each pair of cells intersect. This constraint is repeated for each pair of cells.

    \begin{equation}
         x_{intersect} = \frac{Img(\vec{s}_n - \vec{s}_m)} {tan(\theta_m) - tan(\theta_n)}
        \label{interPt}
    \end{equation}

    \begin{equation}
          x_{intersect} \cap [ (x_{m-1}, x_m) \cup (x_{n-1}, x_n) ] = \emptyset
        \label{interSet}
    \end{equation}
    
    Upon observing (\ref{sum_vectors_eq}) - (\ref{interSet}) they are recognized as a recursive nonlinear system of equations that depend on the previous cell in the series. However, loopy is orientation free; there is no start or end of the series, thus making a closed-form solution difficult to find. Moreover, since Loopy is a decentralized system, these global constraints will not be explicitly checked by the software and will instead be enforced through morphological computation of the inter-cellular links.  
    
 \subsection{Bottom-up Design using Reaction-Diffusion Equations for Morphogenesis}
 \label{section:reactDiff}

     The general reaction-diffusion equation Turing proposed \cite{turing1990chemical} is shown in (\ref{reactDiff}), where $\boldsymbol{Q}$ is a vector that stores the quantity of each simulated chemical, or morphogen, $\boldsymbol{\Gamma}$ is the diagonal matrix of diffusion coefficients, and $R(\boldsymbol{Q})$ is the reaction function that depends on $\boldsymbol{Q}$. 

    \begin{equation}
        \dot{\boldsymbol{Q}} = \boldsymbol{\Gamma} \nabla^2 \boldsymbol{Q} + R(\boldsymbol{Q})
        \label{reactDiff}
    \end{equation}

    Loopy employs a three-morphogen system consisting of an activator ($q_{act}$), an inhibitor ($q_{inh}$), and a passive ($q_{pas}$) morphogen. The passive morphogen only diffuses and does not interact with the other morphogens. Therefore, it encodes a uniform shape (i.e., a circle at steady-state). However, the activator and inhibitor morphogens can react to create stable non-uniform distributions, resulting in the deformation of Loopy's shape. This phenomenon is expressed by \eqref{M2ang}, where the passive and activator morphogens are combined to determine the output angle of each cell.
    
    \begin{equation}
        \theta_m = q_{pas} + q_{act}
        \label{M2ang}
    \end{equation}

    To ensure that the sum of Loopy's joint angles is constant \eqref{polygonSumAngles}, the passive morphogen was initialized with Loopy's starting joint configuration. Next, the Fitzhugh-Nagumo equations were selected as the reaction function because, at steady-state, a sinusoidal distribution is formed with zero mean, as explained in \cite{fitzhugh1955mathematical}. This choice ensures the preservation of the sum of joint angles at steady-state and if the amplitude of the distribution is limited, loopy will not cross and \eqref{interPt} - \eqref{interSet} will be satisfied. 
    
    The reaction-diffusion equations for each of Loopy's morphogens are expressed in (\ref{Qpassive}), (\ref{Qactivator}), and (\ref{Qinhibitor}), where $\alpha$ and $\beta$ are constants that represent the persistent stimulation rate and rate of inhibition, respectively, additionally $\gamma_{pass}$, $\gamma_{act}$, and $\gamma_{inh}$ are the diffusion coefficients of each morphogen.

    \begin{equation}
        \dot{q}_{pas} = \gamma_{pas} \nabla^2 q_{pas}
        \label{Qpassive}
    \end{equation}

    \begin{equation}
        \dot{q}_{act} = \gamma_{act} \nabla^2 q_{act} +  q_{act} - q_{act}^3 - q_{inh} + \alpha
        \label{Qactivator}
    \end{equation}

     \begin{equation}
        \dot{q}_{inh} = \gamma_{inh} \nabla^2 q_{inh} + \beta(q_{act} - q_{inh})
        \label{Qinhibitor}
    \end{equation}

    In Loopy, the communication between cells is restricted to immediate neighbors, meaning that the $N$ finite cells can only transfer morphogens to adjacent cells. Therefore, to model Loopy's behavior accurately, the continuous partial differential Equations of \eqref{Qpassive}, \eqref{Qactivator}, and \eqref{Qinhibitor} are discretized into control volumes of size $s$ (the length of each motor), where the reactions occur and morphogens diffuse across the surfaces. As a result, for the $m^{th}$ cell in the loop, \eqref{Qpassive}, \eqref{Qactivator}, and \eqref{Qinhibitor} reduce to \eqref{Qpd}, \eqref{Qad}, and \eqref{Qid}, respectively. These equations are then explicitly propagated through time ($t$) using \eqref{time_update}, where $\Delta t$ is the time-step.

    \begin{equation}
        \dot{q}_{pas_{m,t}} = \gamma_{pas} \frac{q_{pas_{m-1}} - 2q_{pas_m} + q_{pas_{m+1}}}{2s}
        \label{Qpd}
    \end{equation}

    \begin{equation}
        \begin{aligned}
        \dot{q}_{act_{m,t}} =  \gamma_{act} \frac{q_{act_{m-1}} - 2q_{act_{m}} + q_{act_{m+1}}}{2s} \\ 
        +  q_{act_{m}} - q_{act_{m}}^3 - q_{inh_{m}} + \alpha
        \label{Qad}
        \end{aligned}
    \end{equation}

             \begin{equation}
        \begin{aligned}
        \dot{q}_{inh_{m,t}} =  \gamma_{inh} \frac{q_{inh_{m-1}} - 2q_{inh_{m}} + q_{inh_{m+1}}}{2s} \\
        + \beta(q_{act_{m}} - q_{inh_m})
        \label{Qid}
        \end{aligned}
    \end{equation}

    \begin{equation}
        \boldsymbol{Q}_{m,t+1} =  \dot{\boldsymbol{Q}}_{m,t}\Delta t + \boldsymbol{Q}_{m,t} 
        \label{time_update}
    \end{equation}

%================================================================== RESULTS AND DISCUSSION

\epstopdfDeclareGraphicsRule{.tif}{png}{.png}{convert #1 \OutputFile}
\AppendGraphicsExtensions{.tif}

\section{Experiments, Results $\And$ Discussion}
\label{RaD}
\subsection{Experimental Setup}

Two experiments were conducted to evaluate the emergent shapes of Loopy. The first experiment was conducted in simulation and validated on Loopy to evaluate the impact of each parameter on Loopy's initial steady-state shape and to confirm the validity of the parameters under Loopy's constraints \eqref{sum_vectors_eq}-\eqref{interSet}. The second experiment was conducted to determine the effects on Loopy's final shape when undergoing a step ramp-up-and-down trajectory in parameter space.

To maintain stability \cite{fitzhugh1955mathematical,cevikel2022solitary,gambino2022cross}, the persistent stimulation rate ($\alpha$) was held constant at a small number near zero (.001) in both experiments. The diffusion coefficient of the passive morphogen, $\gamma_{pass}$, was set to a constant value of 50 to allow the primary diffusion process to occur before the reaction effects become dominant. Additionally, to produce non-uniform stable distributions, the inhibitor diffusion rate ($\gamma_{inh}$) must be much larger than the activator diffusion rate ($\gamma_{act}$) \cite{fitzhugh1955mathematical,cevikel2022solitary,gambino2022cross}. Therefore, the diffusion ratio of $\gamma_{inh}$ and $\gamma_{act}$, denoted as $\lambda$, was analyzed to highlight this effect.

\subsection{ Parametric Effects on the Initial Emergent Shape of Loopy}

\begin{figure}[htbp]
\centerline{\includegraphics[scale = .425]{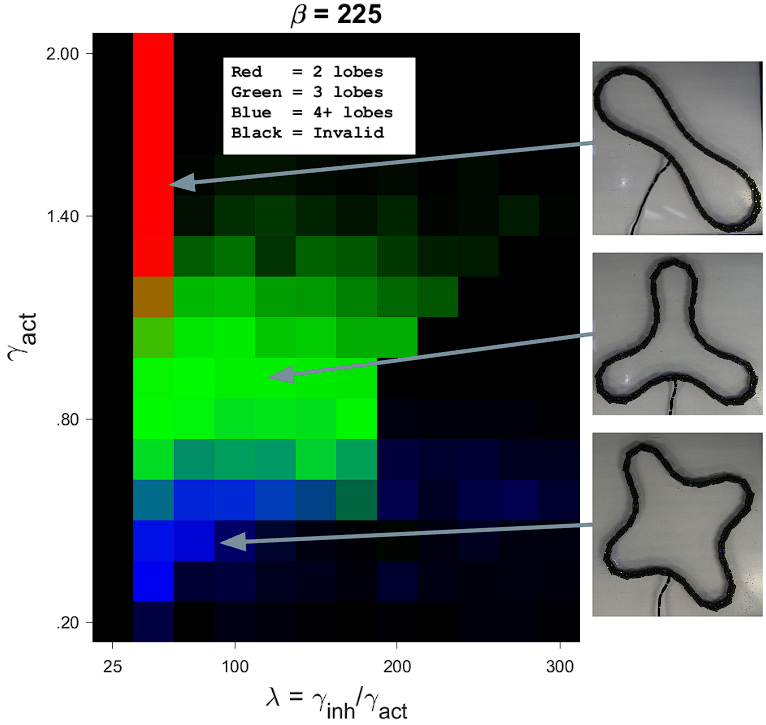}}
\caption{ The parameter-space of the diffusion ratio ($\lambda$) and the activator diffusion rate ($\gamma_{act}$), for the number of lobes Loopy forms when the activator is initialized by small Gaussian noise and with the inhibition rate, $\beta$ = 225. The quantity of red represents the percentage of two lobes that emerged, green is the percentage of three lobes, blue is the percentage of four plus lobes from a set of 50 trials. Additionally, black represents invalid configurations that do not meet Loopy's physical constraints.}
\label{loopylobes}
\end{figure}

Due to the Fitzhugh-Nagumo model generating sinusoidal distributions at steady-state, Loopy develops multiple symmetric protrusions, termed lobes, as shown earlier in Fig. \ref{coolLoopy}. The parametric effects of the activator diffusion rate ($\gamma_{act}$) and the diffusion ratio ($\lambda$) on the steady-state number of lobes are illustrated in Fig. \ref{loopylobes}. The effects of the inhibition rate ($\beta$) on the steady-state lobe count are minimal and thus were held constant at 225. The activator morphogen quantity of each cell was initialized with a small Gaussian noise with zero mean and a standard deviation of .001 with units of morphogens, while the inhibitor was initialized to zero since it is produced by the activator \eqref{Qid}.

%---------------------------------------------------------------------------------------
% parameter space discussion

The first observation of Fig. \ref{loopylobes} is that the number of formed lobes appear in noisy regions with blurred boundaries in the parameter space. This shows that Loopy is able to form multiple emergent shapes with the same parameters (due to the multi-coloring) and is therefore dependent on the initial conditions of the activator morphogen, and even the small amount of noise used to initialize the morphogen can change the number of lobes formed.

Additionally, the figure illustrates that $\gamma_{act}$ governs the number of lobes formed. As $\gamma_{act}$ increases, the number of lobes also increases. Conversely, the parameter $\lambda$ generally had a minor effect on the emergent shape. A potential reason why $\gamma_{act}$ is the dominant parameter is due to its control over the activator's diffusion rate between cells. When the diffusion rate increases, more cells need to be utilized in the reaction portion of \eqref{Qpd}-\eqref{Qid} to balance the heightened diffusion rate. This ultimately limits the total number of lobes on the fixed length of Loopy. 

\begin{figure}[htbp]
\centerline{\includegraphics[scale = .325]{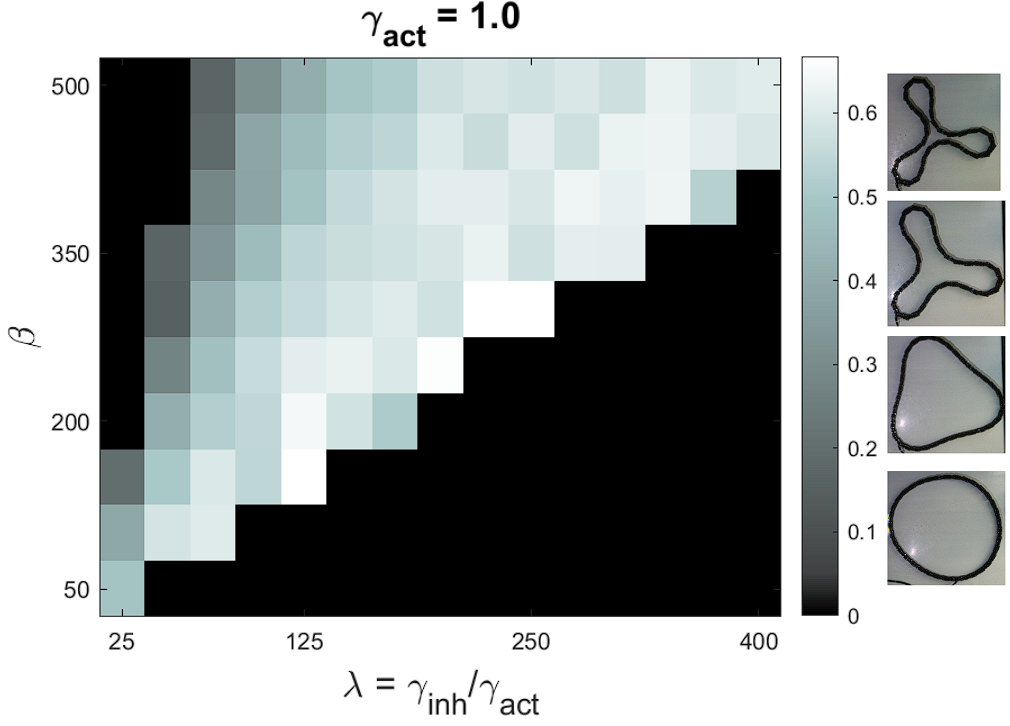}}
\caption{ The parameter-space of the inhibition rate ($\beta$) and the diffusion ratio ($\lambda$), for the amplitude of lobes Loopy forms when the activator is initialized by small Gaussian noise and with the activator diffusion rate, $\gamma_{act}$ = 1. Larger amplitudes are indicated by lighter gray, while black denotes an invalid shape. Note that as the amplitude increases Loopy's shape becomes more pronounced. } 
\label{loopyamplitude}
\end{figure}

%--------------------------------------------------- amplitude
The parametric effect on the average amplitude of the activator morphogen distribution across 50 trials is displayed in Fig. \ref{loopyamplitude}. It is worth noting that the parameter $\gamma_{act}$ has only a minor impact on the final amplitude of the morphogen wave and, as such, was kept constant at 1.

% -----------------------------------------------------------------

Fig. \ref{loopyamplitude} shows that increasing $\beta$ (the rate of inhibition) and decreasing $\lambda$ (the diffusion ratio) reduces the amplitude of the activator morphogen, resulting in nearly circular shapes.  There is a linear boundary for valid configurations, where increasing $\lambda$ requires a corresponding increase in $\beta$ since increasing $\lambda$ or decreasing $\beta$ tends to increase the amplitude, resulting in more pronounced lobes and valleys (as seen to the right of the plot) that can cause Loopy to intersect itself.

%------------------------------------------------ loopy time picture
\begin{figure*}[htbp]
\centerline{\includegraphics[scale = .59]{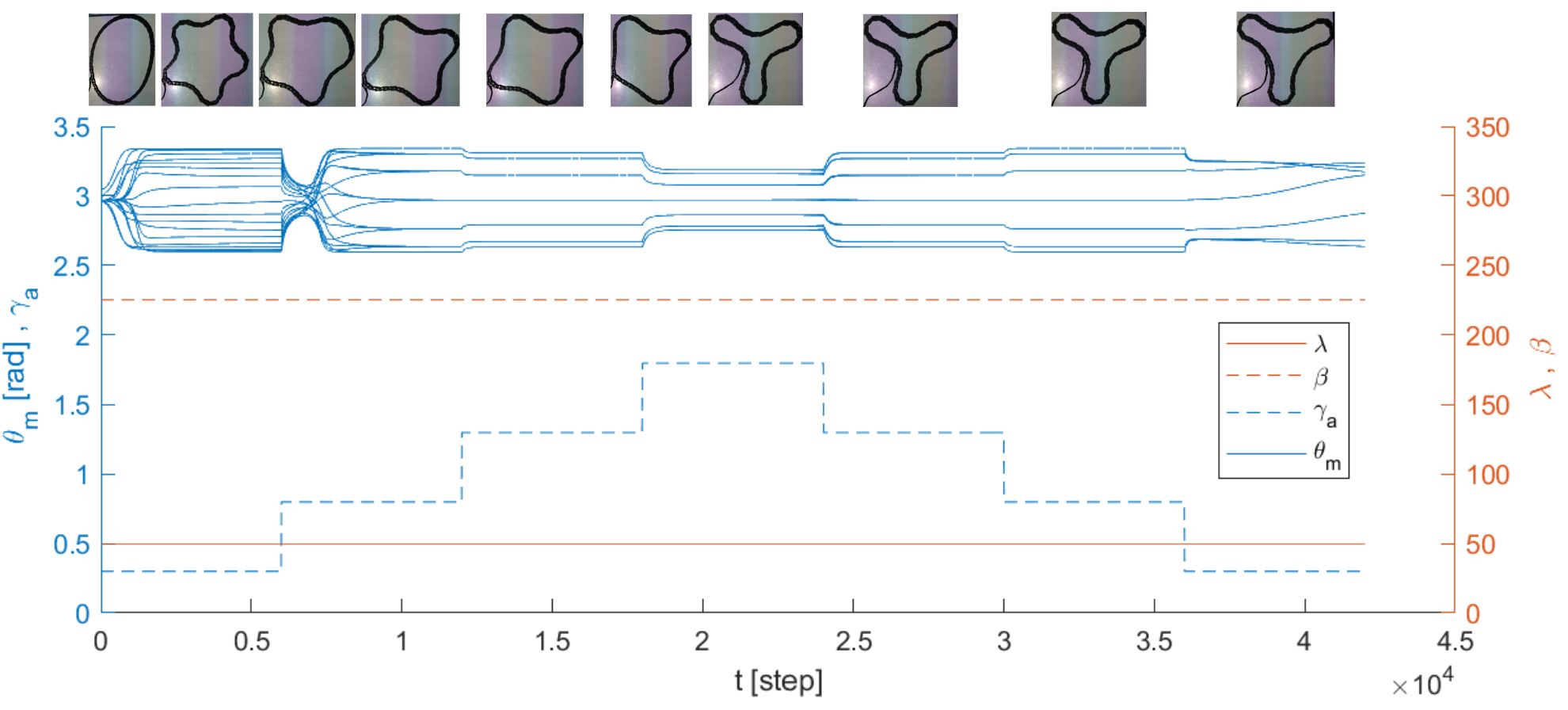}}
\caption{ The transience for each of Loopy's motor responses ($\theta_m$) to a step-ramp-up-and-down parameter trajectory of the activator diffusion coefficient ($\gamma_{act}$) while holding the diffusion ratio ($\lambda$) and  inhibition rate ($\beta$) constant at 50 and 225 respectively. In addition, images of Loopy are displayed at key frames of the trajectory. Note that as Loopy destroys a lobe the amplitude to of the motor angles decreases until the transition is complete, and then it rises to the new equilibrium. Furthermore, as Loopy progresses through the trajectory it is unable to reform the original shape despite the parameters being identical.}
\label{loopytime}
\end{figure*}
% -----------------------------------------------------------------

\subsection{Parameter Trajectory Analysis} % Perturbations 

This experiment was conducted to analyze Loopy's response to run-time changes in parameters. First, as observed in Fig. \ref{loopylobes}, the diffusion rate of the activator ($\gamma_{act}$) is the dominating parameter for the number of emergent lobes Loopy forms. Therefore, to analyze this parameter, the inhibition rate ($\beta$) and the diffusion ratio ($\lambda$) were kept constant at 225 and 50, respectively, while $\gamma_{act}$ was varied to incrementally display each lobe count from five to two lobes. Furthermore, to analyze hysteresis effects, the path was then reversed to return to the starting parameter set. Fig. \ref{loopytime} shows the  motor angles and images of Loopy over time in response to changes in $\gamma_{act}$.

When lobes are formed or destroyed, in Fig. \ref{loopytime}, the amplitude of the morphogens decreases and then increases when the change is complete, as indicated by the pinching of the motor angles ($\theta_m$) curves. This behavior is not caused explicitly by the parameter change, as shown in Fig. \ref{loopyamplitude}, because $\beta$ and $\lambda$ remain constant. Instead, it is an emergent behavior of each cell diffusing its activator morphogens, leading to a decrease in amplitude until one lobe becomes dominant and is removed/added, after which the morphogens can re-accumulate, increasing the amplitude.

Fig. \ref{loopytime} also reveals that when lobes are being destroyed, Loopy is able to destroy them one at a time. However, when the process is reversed, Loopy remains in the three-lobe configuration indicating a hysteresis resistance to forming new lobes. Fig. \ref{loopylobes} supports this for destroying lobes also, as the transition from five lobes to four required a $\gamma_{act}$ value of .8, well within the three-lobed region, and a $\gamma_{act}$ value of 1.8 was needed to transition to three lobes, at the top of the two-lobe region. This resistance is likely due to the morphogens reaching stable equilibria at each lobe frequency, requiring a large change in the parameters to escape these local minima.

\begin{figure}[htbp]
\centerline{\includegraphics[scale = .22]{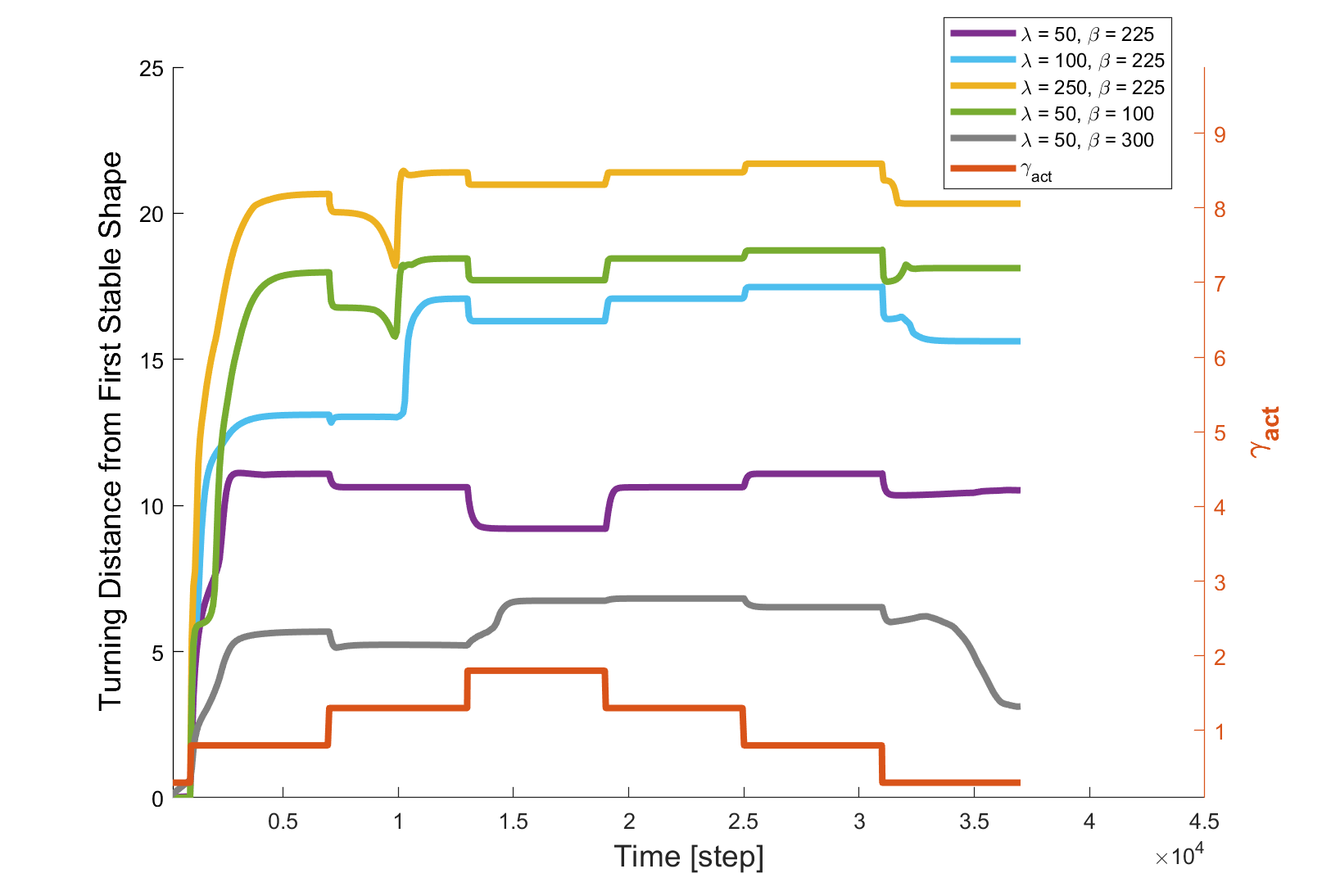}}
\caption{ The turning distance from the starting stable shape as Loopy follows a step-ramp-up-and-down trajectory of the activator diffusion rate ( $\gamma_{act}$) in parameter space for the diffusion ratio, $\lambda = 50, 100, 250$ with the inhibition rate, $\beta = 225$ and $\beta = 100, 225, 300$ with $\lambda = 50$. As $\lambda$ increases, the turning distance from the original shape increases, meaning that the shapes become increasingly different from the original shape. As $\beta$ increases, the turning distance from the original shape decreases, meaning that the shapes remain more similar to the original shape.
}
\label{shape_trans}
\end{figure}

The effect of the diffusion ratio ($\lambda$) and inhibition rate ($\beta$) on the emergent shape was analyzed using the same $\gamma_{act}$ trajectory, with $\lambda$ evaluated at 100 and 250, with $\beta$ held constant at 225, and $\beta$ evaluated at 100 and 300 with $\lambda$ held constant at 50. The turning distance, a polygon-based shape comparison metric \cite{veltkamp2001shape}, was then used to compare the emergent shape trajectories. Fig. \ref{shape_trans} shows the turning distance between the first stable shape and each shape along the parameter trajectories.
%------------------------------------------------ shape comparison plot for Lambda 
% -----------------------------------------------------------------
The first observation from Fig. \ref{shape_trans} is that the final shape is significantly different from the initial shape for each parameter trajectory, further indicating a hysteresis effect when traversing the parameter space.

Another observation from Fig. \ref{shape_trans} is that increasing $\lambda$ results in a greater difference between the final and initial shape, suggesting that the hysteresis resistance is amplified. However, when examining the increase in turning distance at the point where $\gamma_{act}$ first equals 0.8, it becomes apparent that the change in turning distance also increases as $\lambda$ increases. Moreover, as $\lambda$ increases, the small changes in shape observed during the return trajectory when $\lambda = 50$ become less significant. Therefore, the observed hysteresis resistance is better characterized as ``inertia." Loopy is more resistant to changing shape if the shape is not currently changing, but if the shape is in the process of changing, Loopy will continue to change shape.       

To investigate the impact of $\beta$ on the emergent shape, the same $\gamma_{act}$ trajectory was used while keeping $\lambda$ constant at 50, and $\beta$ was evaluated at two additional values: 100 and 300, also shown in Fig. \ref{shape_trans}. Similarly to $\lambda$, the final emergent shapes are significantly different from the initial stable shape. As $\beta$  increases, the magnitude of the turning distance's response to a change in $\gamma_{act}$ decreases. This suggests that $\beta$ has the reverse effect of $\lambda$, where the hysteresis ``inertia" decreases as $\beta$ increases.

\subsection{Discussion on Properties of the Emergent Bodies}
Loopy's emergent morphology displays several interesting features. The most notable is the radial symmetry of each steady-state shape. This symmetry likely arises from the diffusion of morphogens, where any lobe that is larger will diffuse into the other lobes until reaching an equilibrium state \cite{fitzhugh1955mathematical,cevikel2022solitary,gambino2022cross}. Although the steady-state shape of lobes in a system appears to be symmetric, small finite differences of morphogens exist between each lobe. This is because the true steady-state morphogen distribution only occurs as time approaches infinity. These differences play a crucial role in determining where the lobes are destroyed or created with parameter changes \cite{fitzhugh1955mathematical,cevikel2022solitary,gambino2022cross}. For instance, if the lobe count decreases, as shown in Figure \ref{loopytime}, the amplitude of each lobe decreases until one becomes dominant due to the inhibitor reaction \cite{fitzhugh1955mathematical,cevikel2022solitary,gambino2022cross}. Once the dominant lobe is destroyed, the activator reaction becomes dominant, allowing each lobe to regain the lost morphogen stock, as depicted in Figure \ref{loopytime}.   

Another noteworthy observation about emergent bodies is that they exhibit an integer number of periods of morphogens within each shape. This property is enforced by the periodic or closed-loop boundary condition applied to Loopy's cells. As a result, the morphogen concentration in adjacent cells must be both continuous and periodic when traversing around the loop. 

\begin{figure}[htbp]
\centerline{\includegraphics[scale = .35]{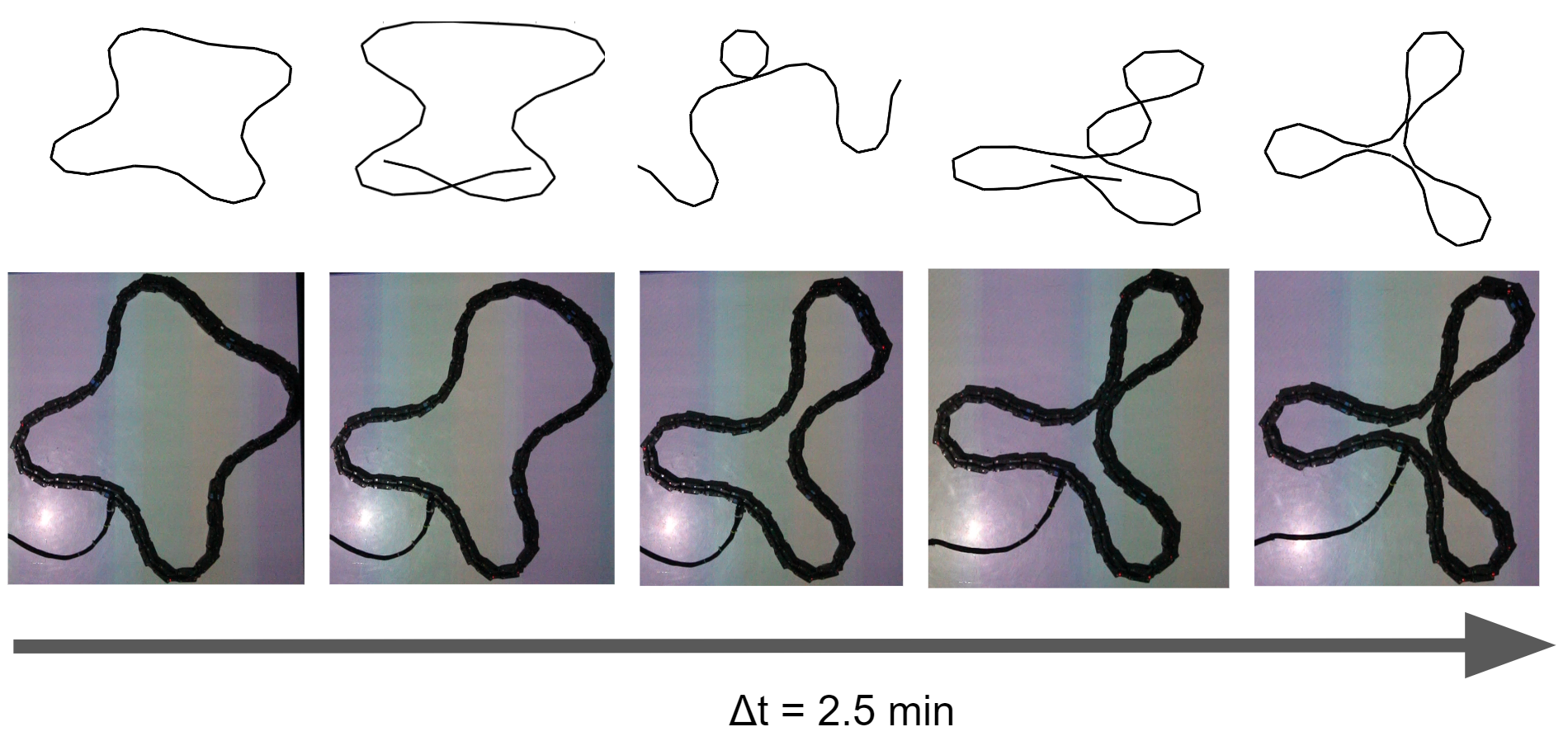}}
\caption{ As Loopy transitions between two valid 
 steady-state configurations, the morphogens often try to enforce drastically invalid configurations (top middle). However, Loopy is able to accommodate these temporary discrepancies through the morphological computation of its inter-cellular links and form the stable valid shape (bottom).}
\label{bad_loopy}
\end{figure}

As Loopy moves through the parameter space, it is not uncommon for significantly invalid configurations to arise, especially during the formation or destruction of a lobe, as shown in Figure \ref{bad_loopy}. However, Loopy can handle these discrepancies through morphological computation, where the error is distributed over the cells. This is possible because the inter-cellular physical constraints, represented by \eqref{sum_vectors_eq}-\eqref{interSet}, are physically enforced rather than explicitly computed. As a result, Loopy's body can correct for temporarily invalid configurations along its parameter trajectory, and still emerge as a symmetric stable body.

%------------------------------------------------ Morphological computation plot 
% -----------------------------------------------------------------

We see Loopy's observed ``inertia" in creating or destroying lobes is a desirable trait for robotic swarms \cite{millonas1994swarms}. This quality helps to conserve energy and minimize the risk of partial failure during transitions caused by small changes in parameters. In a way, Loopy's behavior can be seen as a form of homeostasis that helps to maintain a stable shape equilibrium.

%================================================================== CONCLUSION AND FUTURE WORK
\section{Conclusions $\And$ Future Work}
\label{CaF}
    
   In conclusion, Loopy possesses the ability to generate a diverse range of symmetric stable emergent bodies using a reaction-diffusion system. Once a stable body is formed, Loopy develops an ``inertia" to transfiguration, preventing it from forming or destroying lobes unless there is a significant change in parameters. Moreover, Loopy's morphological computation is capable of correcting temporary and significantly invalid shapes during its trajectory.

    In future work, this study aims to analyze the impact of extracellular environments by introducing a simple sensor to each cell to disrupt the symmetry of emergent bodies and enable them to adapt to their surroundings, similar to how plant roots grow around rocks \cite{li2022plant}. Another area for future study is to explore the simultaneous generation of Loopy's body morphology and behavior (e.g., movement) through a unified bottom-up approach.

\section*{Acknowledgment}
 The authors would like to thank Dylan Covell for his work on the design and construction of the Loopy platform. We would also like to thank Dr. Xi Yu, Dr. Jason Gross, Dr. Guilherme A. S. Pereira, Dr. Ali Baheri, Sarah Alderman, and several participants of West Virginia University’s 2022 robotics Research Experiences for Undergraduates program for insightful discussions.

\bibliographystyle{IEEEtran}
\bibliography{references}
\end{document}